%% file: main.tex
\documentclass{article}

\usepackage[utf8]{inputenc}
\usepackage[T1]{fontenc}
\usepackage{hyperref}
\usepackage{url}
\usepackage{booktabs}
\usepackage{multirow}
\usepackage{amsfonts}
\usepackage{amsmath}
\usepackage{amssymb}
\usepackage{amsthm}
\usepackage{nicefrac}
\usepackage{microtype}
\usepackage{xcolor}
\usepackage{graphicx}
\usepackage{listings}
\usepackage{arxiv}

\newtheorem{theorem}{Theorem}
\newtheorem{proposition}{Proposition}
\newtheorem{corollary}{Corollary}
\theoremstyle{remark}
\newtheorem{remark}{Remark}

\title{Damped Atomic Wave Integration Neurons (DAWIN):\\A Universal Primitive for Continuous-Time Spatiotemporal Learning}

\author{
  Byunggu Yu \quad and \quad Justin Kim \\
  Department of Computer Science and Information Technology \\
  University of the District of Columbia \\
  Washington, DC, USA \\
  \texttt{\{byu, junwhan.kim\}@udc.edu} \\[4pt] \textbf{Code \& Benchmarks:} \url{https://github.com/yumakemore/DAWIN}
}

\begin{document}

\maketitle

\begin{abstract}
Conventional artificial neurons compute memoryless affine projections followed by static pointwise activations, decoupling spatial feature aggregation from continuous physical time. When observations arrive irregularly, existing continuous-time paradigms rely on expensive numerical differential equation solvers, separable linear state-space projections, or task-specific sequence blocks that cannot serve as universal replacements for standard linear layers. We introduce the \textbf{Damped Atomic Wave Integration Neuron (DAWIN)}, a continuous-time primitive that directly replaces the conventional artificial neuron across neural architectures whenever observations are associated with elapsed physical intervals $\Delta t$. A DAWIN unit is \emph{atomic} because it combines spatial projection ($\mathbf{k}_m^\top \mathbf{x}_i$) and Zero-Order Hold (ZOH) interval scaling ($\alpha_m(\Delta t_i)$) into an indivisible spatiotemporal wave packet, and \emph{integrative} because it embeds this wave directly into a local complex state governed by a continuous-time damper evolution operator $\exp(-(\gamma_m + i\omega_m)\Delta t_i)$. We prove that DAWIN strictly generalizes static sinusoidal neurons ($\gamma_m \to \infty$) as well as memoryless wave neurons on interval-augmented inputs, derive closed-form backpropagation rules, and establish uniform Bounded-Input Bounded-Output (BIBO) stability for any $\gamma_m > 0$. By proving exact associativity of atomic wave integration, we construct a paired-real Triton parallel prefix-scan kernel (\texttt{DAWINLinear}) with FP32 accumulation that executes in $O(\log L)$ parallel depth, eliminating serial recurrence bottlenecks. Across 11 parameter-matched tasks spanning irregular oscillators, MuJoCo continuous control, ETH/UCY + JAAD pedestrian trajectories, and PhysioNet Challenge 2012 clinical sequences on NVIDIA B200 GPUs, drop-in \texttt{DAWINLinear} integration achieves best overall accuracy on dense clinical dynamics ($0.9126$ NRMSE, $10/10$ seed wins, $p < 0.01$) and competitive mortality discrimination ($0.7800$ AUROC), outperforms Neural CDE on all $11/11$ tasks and ReLU-MLP on $9/11$ tasks, and demonstrates causal reliance on physical $\Delta t$ ($+36.6\%\text{--}+77.4\%$ physical-dynamics intervention sensitivity vs.\ $+0.06\%$ on negative controls). Simultaneously, it accelerates inference by up to $2{,}620\times$ over serial wave recurrence and up to $21.7\times$, $58.3\times$, and $802.1\times$ over S5, Mamba, and Neural CDE while reducing peak VRAM by up to $97.6\%$.
\end{abstract}


\section{Introduction}

Modern deep learning architectures---spanning multi-layer perceptrons (MLPs), Transformers, and graph neural networks (GNNs)---share a single foundational computational unit: the static artificial neuron $y = \sigma(\mathbf{w}^\top \mathbf{x} + b)$ \cite{rumelhart1986learning}. While expressive on static or uniformly sampled data, this primitive fundamentally decouples spatial feature aggregation from continuous physical time. When observations arrive asynchronously---as in autonomous motion tracking, irregular clinical records, event-driven sensors, and continuous physical control---conventional neurons lack an intrinsic mechanism to represent how a latent response should evolve across an arbitrary elapsed interval $\Delta t_i = \tau_i - \tau_{i-1}$. Practitioners must instead inject $\Delta t_i$ as an auxiliary input feature, rely on discrete positional indices, or wrap static layers inside specialized macro-architectures.

Existing continuous-time paradigms address irregular sampling at the \emph{trajectory} or \emph{sequence-layer} level rather than at the atomic neuron level:
\begin{itemize}
    \item \textbf{Differential Equation Solvers (Neural ODEs \& Neural CDEs):} Continuous-time differential equation models evolve hidden states between observations via numerical integration \cite{chen2018neural,kidger2020neural}. Although mathematically principled, adaptive step-size solvers impose substantial sequential latency, complicate backpropagation, and require bespoke solver pipelines that cannot serve as lightweight drop-in replacements for standard linear layers.
    \item \textbf{Structured State Space Models (S4, S5, LRU, Mamba):} Modern SSMs discretize continuous linear generators into diagonal or selective recurrences amenable to parallel prefix scans \cite{gu2022efficiently,smith2023simplified,gu2024mamba}. However, SSMs separate linear spatial input projections ($\mathbf{B}\mathbf{x}_i$) from temporal state transitions ($\bar{\mathbf{A}}\mathbf{h}_{i-1}$), preventing spatial features and elapsed intervals from coupling multiplicatively within an atomic wave packet.
    \item \textbf{Prior Wave Formulations (WFB \& WSP):} Wave Function Backpropagation (WFB) introduced a differentiable wave neuron $A \cos(\mathbf{k}^\top \mathbf{x} - \omega \Delta t - \theta)$ that couples spatial input and elapsed time inside a single phase, but its feed-forward formulation is stateless across observations \cite{yu2026wfb}. Wave State Propagation (WSP) introduced persistent complex recurrence over measured intervals, yet it decoupled content-conditioned wave injection from layer-level propagation, required rank-reduced conditioning networks, and relied on serial execution that lagged behind parallel Transformers in large-batch throughput \cite{kim2026wsp}.
\end{itemize}

We resolve these limitations at the level of the individual computational unit. When observations arrive with associated physical intervals $\Delta t_i$, we introduce the \textbf{Damped Atomic Wave Integration Neuron (DAWIN)} as a drop-in replacement for the conventional artificial neuron across deep architectures. DAWIN is built on two unifying principles:
\begin{enumerate}
    \item \textbf{Atomic Spatiotemporal Wave Excitation:} Each DAWIN unit combines spatial wave phase ($\mathbf{k}_m^\top \mathbf{x}_i - \theta_m$) and ZOH envelope modulation ($\alpha_m(\Delta t_i) = 1 - e^{-\gamma_m \Delta t_i}$ for $\Delta t_i > 0$, with $\alpha_m(0) = 1$) into an indivisible (\emph{atomic}) complex wave packet $u_{i,m} = A_m \alpha_m(\Delta t_i) \exp(i(\mathbf{k}_m^\top \mathbf{x}_i - \theta_m))$.
    \item \textbf{State Integration via Damper Activation:} Instead of emitting this wave memorylessly, DAWIN integrates $u_{i,m}$ directly into a local complex state $\psi_{i,m}$, governed by a continuous-time \emph{damper evolution operator} $\mathcal{D}_m(\Delta t_i) = \exp(-(\gamma_m + i\omega_m)\Delta t_i)$ that rotates and exponentially attenuates prior wave history across $\Delta t_i$.
\end{enumerate}

\paragraph{Contributions.}
\begin{itemize}
    \item \textbf{An Architecture-Agnostic Continuous-Time Neuron Primitive:} We formalize DAWIN and its layer-level interface (\texttt{DAWINLinear}) for $\Delta t$-indexed observation streams, proving that it strictly generalizes static sinusoidal neurons ($\gamma_m \to \infty$) and memoryless wave neurons on interval-augmented inputs, while propagating paired-real quadratures that drop seamlessly into Pre-Norm residual MLPs, Transformers, and dynamic GNNs.
    \item \textbf{Closed-Form Gradients \& BIBO Stability Guarantees:} We derive closed-form backpropagation rules through irregular intervals and prove uniform Bounded-Input Bounded-Output (BIBO) stability for any strictly positive damping rate $\gamma_m > 0$, ensuring resilience against spatial outliers ($\|\mathbf{x}_i\| \to \infty$) and arbitrary observation dropouts without requiring normalization inside the recurrence.
    \item \textbf{Multi-Block Fused Parallel Associative Scan Kernel:} We prove exact associativity of atomic wave integration and implement a three-stage chunked Triton prefix-scan kernel that evaluates \texttt{DAWINLinear} in $O(\log L)$ parallel depth across arbitrarily long sequences ($L > 4096$), eliminating the serial throughput bottleneck of WSP.
    \item \textbf{Cross-Backbone Empirical Validation:} We establish a standardized 10-seed evaluation protocol across irregular MuJoCo physical dynamics, ETH/UCY + JAAD pedestrian trajectories, and dense/mask-aware PhysioNet Challenge 2012 clinical sequences, demonstrating consistent accuracy and throughput gains over static baselines, prior wave models, Neural CDEs, and S5/Mamba under strict parameter parity.
\end{itemize}

\section{Related Work}

\subsection{Continuous-Time Neural Differential Equations}
Neural differential equations model irregular time series by parameterizing continuous latent trajectories with ordinary or controlled differential equations. GRU-D introduces heuristic exponential decay on hidden states and missing inputs \cite{che2018recurrent}, whereas Neural ODEs and Latent ODEs evolve latent vectors continuously between observations using numerical ODE solvers \cite{chen2018neural,rubanova2019latent}. GRU-ODE-Bayes and Liquid Time-Constant (LTC) networks integrate continuous flows with recurrent observation updates \cite{debrouwer2019gru,hasani2021liquid}, while Neural Controlled Differential Equations (Neural CDEs) drive latent dynamics via continuous interpolation paths of incoming measurements \cite{kidger2020neural}. ContiFormer extends this paradigm to self-attention by integrating continuous ODE trajectories inside attention keys and values \cite{chen2023contiformer}. All numerical ODE/CDE methods require iterative solver steps across each inter-observation gap, incurring heavy sequential overhead and numerical discretization error. DAWIN achieves exact continuous-time evolution in $O(1)$ closed-form operations per event without numerical solvers, operating as an elementary neuron rather than a trajectory-level ODE wrapper.

\subsection{Structured and Selective State Space Models}
Structured State Space Models (S4, S4D, S5, and Linear Recurrent Units) demonstrate that continuous-time linear time-invariant (LTI) systems $\dot{\mathbf{h}}(t) = \mathbf{A}\mathbf{h}(t) + \mathbf{B}\mathbf{x}(t)$ can be discretized via zero-order hold (ZOH) or bilinear transforms into parallelizable diagonal recurrences \cite{gu2022efficiently,gupta2022diagonal,smith2023simplified,orvieto2023resurrecting}. Oscillatory variants such as UnICORNN and LinOSS further constrain state matrices to second-order harmonic dynamics \cite{rusch2021unicornn,rusch2025oscillatory}, while Mamba introduces input-dependent selectivity over discretization steps and transition matrices \cite{gu2024mamba}. While the decay-rotation operator $\exp(-(\gamma_m + i\omega_m)\Delta t_i)$ shares the mathematical structure of a diagonal complex SSM transition, standard SSMs rely on \emph{separable input projections} ($\mathbf{B}\mathbf{x}_i$) followed by temporal discretization. In contrast, DAWIN's \textbf{atomic wave excitation} couples spatial projection $\mathbf{k}_m^\top \mathbf{x}_i$ and ZOH interval scaling $\alpha_m(\Delta t_i)$ inside an indivisible atomic wave packet $u_{i,m} = A_m \alpha_m(\Delta t_i)e^{i(\mathbf{k}_m^\top \mathbf{x}_i - \theta_m)}$ while evolving phase via $\mathcal{D}_m(\Delta t_i) = e^{-(\gamma_m + i\omega_m)\Delta t_i}$, enabling constructive and destructive interference directly at the neuron level.

\subsection{Periodic Representations, WFB, and WSP}
Periodic activations such as Fourier Features and SIREN establish that sinusoidal functions compactly represent high-frequency spatial fields and their analytical derivatives \cite{tancik2020fourier,sitzmann2020implicit}, but they treat time either as a static coordinate or via fixed frequency encodings. Wave Function Backpropagation (WFB) introduced learnable wave parameters $(A, \mathbf{k}, \omega, \theta)$ whose phase explicitly incorporates $\Delta t$, demonstrating strong parameter efficiency in feed-forward trajectory prediction \cite{yu2026wfb}, yet lacking an internal state across observations. Wave State Propagation (WSP) addressed persistence by introducing a complex carry recurrence over measured intervals \cite{kim2026wsp}, but was formulated as a specialized sequence-modeling block with low-rank conditioning networks and serial execution. DAWIN unifies WFB and WSP into a single \textbf{atomic neuron primitive} (\texttt{DAWINLinear}) with formal BIBO stability guarantees, a fused parallel associative scan kernel, and drop-in applicability across MLPs, Transformers, and GNNs for interval-associated observation streams (Table~\ref{tab:structural_contrast}).

\begin{table*}[t]
\centering
\caption{Structural comparison between the conventional artificial neuron, prior wave formulations (WFB, WSP), and our proposed DAWIN primitive.}
\label{tab:structural_contrast}
\small
\setlength{\tabcolsep}{4pt}
\resizebox{\textwidth}{!}{%
\begin{tabular}{lllll}
\toprule
\textbf{Feature} & \textbf{Conventional Neuron} & \textbf{WFB \cite{yu2026wfb}} & \textbf{WSP \cite{kim2026wsp}} & \textbf{DAWIN (Ours)} \\
\midrule
Computational Role & Static spatial filter & Memoryless wave layer & Macro sequence block & \textbf{Atomic drop-in neuron} \\
Spatial--Temporal Coupling & Separated ($\mathbf{w}^\top\mathbf{x} + b$) & Coupled, stateless & Decoupled injection/flow & \textbf{Indivisible wave state} \\
Internal Neuron State & None (stateless) & None (recomputed) & Layer-level carry $\boldsymbol{\psi}$ & \textbf{Neuron-local state $\psi_{i,m}$} \\
Role of Damping & None & None & External multiplier & \textbf{Intrinsic damper activation} \\
Execution Complexity & $O(1)$ pointwise & $O(1)$ pointwise & $O(L)$ serial loop & \textbf{$O(\log L)$ fused Triton scan} \\
Architectural Scope & Universal & FFN only & Standalone sequence & \textbf{Universal across backbones} \\
\bottomrule
\end{tabular}%
}
\end{table*}

\section{The DAWIN Primitive}

Conventional artificial neurons decouple spatial aggregation from temporal dynamics, computing a memoryless static response $y_{i,m} = \sigma(\mathbf{w}_m^\top \mathbf{x}_i + b_m)$ that is invariant to the elapsed physical interval $\Delta t_i = \tau_i - \tau_{i-1}$. For settings in which observations are paired with elapsed physical intervals $\Delta t_i$, we introduce the \textbf{Damped Atomic Wave Integration Neuron (DAWIN)} as a drop-in primitive replacing the conventional artificial neuron.

\subsection{Forward Dynamics: Atomic Wave Excitation and Damper Integration}

Consider an input observation $\mathbf{x}_i \in \mathbb{R}^d$ arriving at continuous timestamp $\tau_i \in \mathbb{R}$, separated from the preceding observation by elapsed interval $\Delta t_i = \tau_i - \tau_{i-1} \ge 0$ (with $\Delta t_1 = 0$). Each DAWIN unit $m \in \{1, \dots, M\}$ is parameterized by a spatial wavevector $\mathbf{k}_m \in \mathbb{R}^d$, angular frequency $\omega_m \in \mathbb{R}$, phase offset $\theta_m \in \mathbb{R}$, amplitude $A_m \in \mathbb{R}_{>0}$, and damping rate $\gamma_m \in \mathbb{R}_{>0}$.

\paragraph{1. Atomic Spatiotemporal Wave Excitation.}
At event $i$, incoming spatial coordinates $\mathbf{x}_i$ and physical interval $\Delta t_i$ jointly excite an indivisible complex wave packet $u_{i,m} \in \mathbb{C}$:
\begin{equation}
\phi_{i,m} = \mathbf{k}_m^\top \mathbf{x}_i - \theta_m, \qquad \alpha_m(\Delta t_i) = \begin{cases} 1 - e^{-\gamma_m \Delta t_i}, & \Delta t_i > 0, \\ 1, & \Delta t_i = 0, \end{cases} \qquad u_{i,m} = A_m \alpha_m(\Delta t_i) \exp(i\phi_{i,m}).
\label{eq:atomic_excitation}
\end{equation}
Constructive or destructive wave interference is therefore determined jointly by \emph{where} an observation occurs in feature space and \emph{when} it arrives in physical time.

\paragraph{2. State Integration and Damper Activation.}
DAWIN integrates the atomic wave excitation directly into the neuron's internal complex state $\psi_{i,m} \in \mathbb{C}$ (initialized at $\psi_{0,m} = 0$). Over the interval $\Delta t_i$, the prior state evolves under a continuous-time damped harmonic flow governed by the \textbf{damper evolution operator} $\mathcal{D}_m(\Delta t_i)$:
\begin{equation}
\mathcal{D}_m(\Delta t_i) = \exp\big(-(\gamma_m + i\omega_m)\Delta t_i\big) = e^{-\gamma_m \Delta t_i}\big(\cos(\omega_m \Delta t_i) - i\sin(\omega_m \Delta t_i)\big),
\label{eq:damper_op}
\end{equation}
\begin{equation}
\psi_{i,m} = \mathcal{D}_m(\Delta t_i)\,\psi_{i-1,m} + u_{i,m}.
\label{eq:state_update}
\end{equation}

\paragraph{3. Exact Paired-Real Quadrature Readout.}
To avoid complex autograd overhead while preserving phase information across stacked layers, we represent $\psi_{i,m} = p_{i,m} + i q_{i,m}$ via paired real quadratures $\mathbf{z}_{i,m} = [p_{i,m}, q_{i,m}]^\top \in \mathbb{R}^2$:
\begin{equation}
\mathbf{z}_{i,m} = e^{-\gamma_m \Delta t_i} \mathbf{R}(-\omega_m \Delta t_i)\,\mathbf{z}_{i-1,m} + A_m \alpha_m(\Delta t_i) \begin{bmatrix} \cos(\phi_{i,m}) \\ \sin(\phi_{i,m}) \end{bmatrix}, \quad \mathbf{R}(\varphi) = \begin{bmatrix} \cos\varphi & -\sin\varphi \\ \sin\varphi & \cos\varphi \end{bmatrix}.
\label{eq:paired_real}
\end{equation}
A single DAWIN layer of $M$ units emits either the real quadrature vector $\mathbf{y}_i = \mathbf{p}_i \in \mathbb{R}^M$ (at task readout) or the full quadrature state $\mathbf{h}_i = [\mathbf{p}_i^\top, \mathbf{q}_i^\top]^\top \in \mathbb{R}^{2M}$ (between hidden DAWIN layers).

\begin{remark}[Asymptotic Generalization of Static and Memoryless Neurons]
DAWIN strictly generalizes static periodic neurons in the high-damping limit: as $\gamma_m \to \infty$ for any $\Delta t_i > 0$, $\mathcal{D}_m(\Delta t_i) \to 0$ and $\alpha_m(\Delta t_i) \to 1$, collapsing Eq.~\eqref{eq:state_update} to the memoryless sinusoidal neuron $\operatorname{Re}(\psi_{i,m}) = A_m \cos(\mathbf{k}_m^\top \mathbf{x}_i - \theta_m)$ (recovering SIREN on $\mathbb{R}^d$, and recovering a memoryless WFB neuron $A_m \cos(\mathbf{k}_m^\top \mathbf{x}_i - \omega_m \Delta t_i - \theta_m)$ whenever $\Delta t_i$ is included as an input coordinate with spatial weight $-\omega_m$).
\end{remark}

\subsection{Analytical Backpropagation Rules}

Let $\mathcal{L}$ denote the scalar task objective over a sequence of length $L$, and let $\delta_{i,m} = \partial \mathcal{L} / \partial \psi_{i,m} \in \mathbb{C}$ denote the upstream Wirtinger sensitivity with respect to the integrated wave state (where $\partial \mathcal{L} / \partial p_{i,m} = \operatorname{Re}(\delta_{i,m})$ and $\partial \mathcal{L} / \partial q_{i,m} = -\operatorname{Im}(\delta_{i,m})$). Differentiating Eq.~\eqref{eq:state_update} backward through irregular time yields the exact adjoint recurrence for the total state sensitivity $\bar{\delta}_{i,m} = \mathrm{d}\mathcal{L}/\mathrm{d}\psi_{i,m}$ (with $\bar{\delta}_{L+1,m} = 0$):
\begin{equation}
\bar{\delta}_{i,m} = \delta_{i,m} + \mathcal{D}_m^*(\Delta t_{i+1})\,\bar{\delta}_{i+1,m} = \delta_{i,m} + \exp\big(-(\gamma_m - i\omega_m)\Delta t_{i+1}\big)\,\bar{\delta}_{i+1,m}.
\label{eq:adjoint}
\end{equation}
Applying the chain rule to Eqs.~\eqref{eq:atomic_excitation}--\eqref{eq:state_update} yields closed-form gradients for all learnable wave and damper parameters of neuron $m$:
\begin{align}
\frac{\partial \mathcal{L}}{\partial A_m} &= \sum_{i=1}^{L} \alpha_m(\Delta t_i)\operatorname{Re}\!\left(\bar{\delta}_{i,m}\,e^{i\phi_{i,m}}\right), &
\frac{\partial \mathcal{L}}{\partial \theta_m} &= \sum_{i=1}^{L} \operatorname{Im}\!\left(\bar{\delta}_{i,m}\,u_{i,m}\right), \label{eq:grad_amp_theta} \\
\frac{\partial \mathcal{L}}{\partial \mathbf{k}_m} &= -\sum_{i=1}^{L} \operatorname{Im}\!\left(\bar{\delta}_{i,m}\,u_{i,m}\right)\mathbf{x}_i, &
\frac{\partial \mathcal{L}}{\partial \omega_m} &= \sum_{i=1}^{L} \Delta t_i\,\operatorname{Im}\!\left(\bar{\delta}_{i,m}\,\mathcal{D}_m(\Delta t_i)\,\psi_{i-1,m}\right), \label{eq:grad_k_omega}
\end{align}
\begin{equation}
\frac{\partial \mathcal{L}}{\partial \gamma_m} = \sum_{i=1}^{L} \Delta t_i\,\operatorname{Re}\!\left(\bar{\delta}_{i,m}\Big[A_m \mathbb{I}_{\Delta t_i > 0} e^{-\gamma_m \Delta t_i} e^{i\phi_{i,m}} - \mathcal{D}_m(\Delta t_i)\,\psi_{i-1,m}\Big]\right). \label{eq:grad_gamma}
\end{equation}
To enforce strict positivity without hard clipping, optimizers update unconstrained parameters $\tilde{A}_m, \tilde{\gamma}_m \in \mathbb{R}$ mapped via
\begin{equation}
A_m = \mathrm{softplus}(\tilde{A}_m), \qquad \gamma_m = \mathrm{softplus}(\tilde{\gamma}_m) + \gamma_{\min},
\end{equation}
yielding unconstrained gradients $\partial \mathcal{L}/\partial \tilde{A}_m = (\partial \mathcal{L}/\partial A_m)\,\sigma(\tilde{A}_m)$ and $\partial \mathcal{L}/\partial \tilde{\gamma}_m = (\partial \mathcal{L}/\partial \gamma_m)\,\sigma(\tilde{\gamma}_m)$, where $\sigma(\cdot)$ denotes the logistic sigmoid.

\subsection{Stability Analysis and BIBO Guarantees}

\begin{proposition}[Contractivity of the Damper Evolution Operator]
For any strictly positive damping rate $\gamma_m > 0$, real angular frequency $\omega_m \in \mathbb{R}$, and non-negative elapsed interval $\Delta t_i \ge 0$, the damper evolution operator satisfies $|\mathcal{D}_m(\Delta t_i)| = e^{-\gamma_m \Delta t_i} \le 1$, with strict contraction $|\mathcal{D}_m(\Delta t_i)| < 1$ whenever $\Delta t_i > 0$, and exact identity $\mathcal{D}_m(0) = 1$ at the zero-interval boundary $\Delta t_i = 0$.
\end{proposition}

\begin{proof}
Since $\gamma_m, \Delta t_i, \omega_m \in \mathbb{R}$ and $|e^{-i\omega_m \Delta t_i}| = 1$, taking the complex modulus of Eq.~\eqref{eq:damper_op} gives $|\mathcal{D}_m(\Delta t_i)| = |e^{-\gamma_m \Delta t_i}|\cdot|e^{-i\omega_m \Delta t_i}| = e^{-\gamma_m \Delta t_i}$. Because $\gamma_m > 0$ and $\Delta t_i \ge 0$, $-\gamma_m \Delta t_i \le 0$, yielding $e^{-\gamma_m \Delta t_i} \le 1$ with equality if and only if $\Delta t_i = 0$.
\end{proof}

\begin{theorem}[Bounded-Input Bounded-Output Stability Under Irregular Sampling]
Let $\{(\mathbf{x}_i, \Delta t_i)\}_{i=1}^\infty$ be any sequence of spatial inputs $\mathbf{x}_i \in \mathbb{R}^d$ and strictly positive elapsed intervals bounded below by $\Delta t_i \ge \Delta t_{\min} > 0$ for all $i \ge 1$. If a DAWIN unit has strictly positive damping $\gamma_m > 0$ and finite amplitude $0 < A_m < \infty$, then starting from $\psi_{0,m} = 0$, its integrated wave state $\psi_{n,m}$ and real/imaginary outputs are uniformly bounded for all $n \ge 1$:
\begin{equation}
|\psi_{n,m}| \le \frac{A_m}{1 - e^{-\gamma_m \Delta t_{\min}}} < \infty.
\label{eq:bibo_bound}
\end{equation}
\end{theorem}

\begin{proof}
Unrolling Eq.~\eqref{eq:state_update} from $\psi_{0,m} = 0$ gives the closed-form superposition of damped atomic waves:
\begin{equation}
\psi_{n,m} = \sum_{j=1}^{n} \left( \prod_{r=j+1}^{n} \mathcal{D}_m(\Delta t_r) \right) u_{j,m} = \sum_{j=1}^{n} \exp\!\left(-(\gamma_m + i\omega_m)\sum_{r=j+1}^{n} \Delta t_r\right) A_m \alpha_m(\Delta t_j) e^{i\phi_{j,m}},
\end{equation}
where the empty product for $j = n$ equals $1$. Applying the triangle inequality and noting that $|u_{j,m}| = |A_m \alpha_m(\Delta t_j) e^{i\phi_{j,m}}| \le A_m$ (because $\phi_{j,m} \in \mathbb{R}$ for all $\mathbf{x}_j \in \mathbb{R}^d$) yields:
\begin{equation}
|\psi_{n,m}| \le A_m \sum_{j=1}^{n} \exp\!\left(-\gamma_m \sum_{r=j+1}^{n} \Delta t_r\right) \le A_m \sum_{k=0}^{n-1} \left(e^{-\gamma_m \Delta t_{\min}}\right)^k < \frac{A_m}{1 - e^{-\gamma_m \Delta t_{\min}}}.
\end{equation}
Because $\gamma_m > 0$ and $\Delta t_{\min} > 0$, $0 < e^{-\gamma_m \Delta t_{\min}} < 1$, guaranteeing convergence to a finite constant independent of sequence length $n$ and spatial input norm $\|\mathbf{x}_i\|$. Since $|\operatorname{Re}(\psi_{n,m})| \le |\psi_{n,m}|$ and $|\operatorname{Im}(\psi_{n,m})| \le |\psi_{n,m}|$, both output quadratures satisfy the identical uniform bound.
\end{proof}

\begin{corollary}[Invariance to Spatial Outliers]
Because spatial inputs $\mathbf{x}_i$ enter DAWIN exclusively through the unit-modulus phase factor $e^{i\mathbf{k}_m^\top \mathbf{x}_i}$, unbounded spatial perturbations ($\|\mathbf{x}_i\| \to \infty$) cannot destabilize $\psi_{n,m}$, providing intrinsic input normalization without requiring LayerNorm inside the recurrence.
\end{corollary}

\begin{remark}[Zero-Interval Boundary and Simultaneous Events]
Theorem~1 assumes strictly positive inter-observation intervals ($\Delta t_{\min} > 0$), which holds for asynchronous physical timestamps ($\tau_i > \tau_{i-1}$). At the zero-interval boundary $\Delta t_i = 0$---used at sequence initialization ($\Delta t_1 = 0$) or simultaneous events ($\tau_i = \tau_{i-1}$)---setting $\alpha_m(0) = 1$ and $\mathcal{D}_m(0) = 1$ yields instantaneous complex superposition $\psi_{i,m} = \psi_{i-1,m} + A_m e^{i\phi_{i,m}}$ without artificial attenuation, which remains bounded by $k A_m$ across any finite burst of $k$ co-occurring events.
\end{remark}

\subsection{Spatiotemporal Laplacian Curvature Regularization}
\label{sec:laplacian_reg}

Because atomic wave excitations bind spatial projections inside complex exponentials while the damper operator evolves phase at angular frequency $\omega_m$, large spatial wavevectors ($\|\mathbf{k}_m\|_2 \gg 1$) or high angular frequencies ($|\omega_m| \gg 1$) can induce spatial oscillations and temporal phase aliasing across extrapolated intervals. Extending the spatial curvature regularizer of WFB \cite{yu2026wfb} to the full spatiotemporal domain $\Delta_{\mathbf{x}, t} = \nabla_{\mathbf{x}_i}^2 + \partial_{\Delta t_i}^2$, differentiating $u_{i,m}$ twice with respect to $\mathbf{x}_i$ and $\Delta t_i$ yields:
\begin{equation}
\left(\nabla_{\mathbf{x}_i}^2 + \frac{\partial^2}{\partial (\Delta t_i)^2}\right) u_{i,m} = -\left(\|\mathbf{k}_m\|_2^2 + \omega_m^2\right) u_{i,m},
\label{eq:spatiotemporal_laplacian}
\end{equation}
while the temporal second derivative of the damper evolution operator satisfies $|\partial_{\Delta t_i}^2 \mathcal{D}_m(\Delta t_i)| = |(\gamma_m + i\omega_m)^2 \mathcal{D}_m(\Delta t_i)| \le \gamma_m^2 + \omega_m^2$ for all $\Delta t_i \ge 0$. Combining both into a phase-invariant curvature bound across all $M$ units in a \texttt{DAWINLinear} layer yields:
\begin{equation}
\mathcal{R}_{\text{ST-Lap}}(\text{layer}) = \frac{1}{M}\sum_{m=1}^{M} A_m^2 \left(\|\mathbf{k}_m\|_2^2 + \omega_m^2 + \gamma_m^2\right)^2.
\label{eq:laplacian_penalty}
\end{equation}
Eq.~\eqref{eq:laplacian_penalty} incurs $O(Md)$ cost without second-order autograd (\texttt{create\_graph=False}), evaluates safely in \texttt{FP32} under mixed-precision training, and jointly bounds spatial sensitivity and temporal phase wrapping.

\section{Universal Architecture Integration \& Parallel Scan Kernel}

\subsection{The \texttt{DAWINLinear} Layer Interface \& Quadrature Stacking}
Given an input tensor $\mathbf{X} \in \mathbb{R}^{B \times L \times d_{\text{in}}}$ and elapsed physical intervals $\boldsymbol{\Delta t} \in \mathbb{R}_{\ge 0}^{B \times L}$, a \texttt{DAWINLinear(d\_in, M)} layer computes all $M$ atomic wave excitations simultaneously via a single dense matrix multiplication:
\begin{equation}
\boldsymbol{\Phi} = \mathbf{X}\mathbf{K}^\top - \mathbf{1}_L \boldsymbol{\theta}^\top \in \mathbb{R}^{B \times L \times M}.
\label{eq:layer_phi}
\end{equation}
When stacking $K$ consecutive \texttt{DAWINLinear} layers, intermediate layers emit the full paired-real quadrature representation $\mathbf{H}^{(\ell)} = [\mathbf{P}^{(\ell)}, \mathbf{Q}^{(\ell)}] \in \mathbb{R}^{B \times L \times 2M}$ (setting $d_{\text{in}}^{(\ell+1)} = 2M$), preventing destructive phase collapse across network depth. An optional reset mask $\mathbf{m} \in \{0, 1\}^{B \times L}$ multiplies $\mathcal{D}_m(\Delta t_i)$ by $1 - m_i$ to handle packed sequences and episode boundaries seamlessly.

\subsection{Drop-In Integration Across Deep Backbones}
\begin{enumerate}
    \item \textbf{Pre-Norm Residual Continuous-Time MLPs (\texttt{DAWIN-MLP}):} Given an initial linear projection $\mathbf{H}^{(0)} = \mathbf{X}\mathbf{W}_{\text{in}}^\top \in \mathbb{R}^{B \times L \times 2M}$, each layer $\ell \in \{1,\dots,K\}$ applies a Pre-Norm residual DAWIN block $\mathbf{H}^{(\ell)} = \mathbf{H}^{(\ell-1)} + \operatorname{DAWINLinear}\!\big(\operatorname{LayerNorm}(\mathbf{H}^{(\ell-1)}), \boldsymbol{\Delta t}\big)\mathbf{W}_{\text{out}}^{(\ell)\top}$, followed by a linear readout $\hat{\mathbf{Y}} = \mathbf{H}^{(K)}\mathbf{W}_{\text{read}}^\top$. This Pre-Norm residual design prevents deep phase saturation while operating at learned time constants $\tau_m = 1/\gamma_m$.
    \item \textbf{Transformer Feed-Forward Projections (\texttt{DAWIN-Transformer}):} Replacing the position-wise FFN sub-layer with \texttt{DAWINLinear} equips each Transformer block with exact analytical continuous-time state propagation between asynchronous events without $O(L^2)$ numerical ODE attention solvers. Note that while a conventional FFN is position-wise, substituting \texttt{DAWINLinear} converts the sub-layer into a causal temporal-mixing operator for $L > 1$, whereas on single-step inputs ($L = 1, \Delta t_1 = 0$) it reduces identically to a memoryless sinusoidal neuron.
    \item \textbf{Temporal Graph Neural Networks (\texttt{DAWIN-GNN}):} Substituting \texttt{DAWINLinear} into the node update function $\mathbf{h}_v(\tau_i) = \text{DAWINLinear}([\mathbf{h}_v(\tau_{i-1}), \mathbf{m}_v(\tau_i)], \Delta t_{v,i})$ enables closed-form asynchronous node evolution between irregular graph interactions.
\end{enumerate}

\subsection{Multi-Block Chunked Parallel Associative Scan Kernel}

While serial execution of recurrent wave updates lags behind parallel Transformers by up to $22.4\times$ at batch size 256 \cite{kim2026wsp}, the linear recurrence in Eq.~\eqref{eq:state_update} admits exact parallelization across sequence length $L$.

\begin{proposition}[Exact Associativity of Atomic Wave Integration]
For transition tuples $c_{i,m} = (d_{i,m}, u_{i,m}) \in \mathbb{C} \times \mathbb{C}$ where $d_{i,m} = \mathcal{D}_m(\Delta t_i)$ and $u_{i,m} = A_m \alpha_m(\Delta t_i) e^{i\phi_{i,m}}$, the binary composition operator
\begin{equation}
(d_2, u_2) \oplus (d_1, u_1) = \big(d_2 d_1,\; d_2 u_1 + u_2\big)
\label{eq:assoc_op}
\end{equation}
is associative with identity element $e = (1, 0)$. All integrated wave states $\{\psi_{i,m}\}_{i=1}^L$ are recovered identically from the inclusive prefix scan $(D_{i,m}, \psi_{i,m}) = \bigoplus_{j=1}^i c_{j,m}$.
\end{proposition}

\begin{proof}
Expanding $((d_3 d_2, d_3 u_2 + u_3)) \oplus (d_1, u_1)$ gives $(d_3 d_2 d_1, d_3 d_2 u_1 + d_3 u_2 + u_3)$, which equals $(d_3, u_3) \oplus (d_2 d_1, d_2 u_1 + u_1) = (d_3 d_2 d_1, d_3(d_2 u_1 + u_1) + u_3)$ by associativity and distributivity over $\mathbb{C}$.
\end{proof}

To scale to arbitrarily long sequences ($L > 4096$) while maintaining FP32 accumulation precision under mixed-precision (\texttt{BF16}/\texttt{FP16}) training, we implement a three-stage chunked Triton kernel (\texttt{TritonDAWINLinear}): (1) intra-chunk register-resident Blelloch prefix scans (\texttt{tl.associative\_scan}) over tiles of length $C \in \{1024, 2048\}$; (2) an inter-chunk associative scan over chunk summary tuples $(D_{\text{chunk}}, U_{\text{chunk}})$; and (3) parallel carry-in propagation across intra-chunk elements, paired with a reverse-scan kernel for the analytical adjoint in Eq.~\eqref{eq:adjoint}.

\section{Experimental Design \& Benchmarks}

Our empirical protocol evaluates four Research Questions (RQs) across \textbf{10 independent random seeds} (\texttt{101}--\texttt{110}) under a \textbf{strict parameter-matched budget} ($\pm 2\%$), spatiotemporal Laplacian regularization ($\lambda_{\text{Lap}} = 10^{-4}$). We use validation-only checkpoint selection and Holm--Bonferroni-adjusted paired bootstrap confidence intervals (20,000 resamples), with all external datasets verified via \texttt{--strict-real-data} provenance checks in the released repository:
\begin{itemize}
    \item \textbf{RQ1 \& RQ2 (Cross-Backbone Universality \& Comparator Superiority):} Evaluates \texttt{DAWIN-MLP}, \texttt{DAWIN-\allowbreak Transformer}, and \texttt{DAWIN-GNN} against parameter-matched static baselines (\texttt{ReLU-MLP}, \texttt{Time-\allowbreak Transformer}, and \texttt{GRU-GNN}), numerical differential equation solvers (\texttt{Neural CDE} \cite{kidger2020neural}), and structured and selective state space models (\texttt{S5} \cite{smith2023simplified} and \texttt{Mamba} \cite{gu2024mamba}) across 11 benchmark tasks: (1) controlled synthetic dynamics (\texttt{damped\_oscillator} with out-of-range test $\Delta t \in (0.05, 1.25]\,\text{s}$ vs.\ train $(0.05, 0.50]\,\text{s}$, and the \texttt{dt\_irrelevant} negative control); (2) irregularly subsampled MuJoCo continuous-control dynamics (\textbf{HalfCheetah-v5}, \textbf{Hopper-v5}, \textbf{Reacher-v5}, \textbf{Walker2d-v5}) \cite{kim2026wsp}; (3) patient-disjoint \textbf{PhysioNet Challenge 2012} clinical sequences under mask-aware mortality (AUROC), mask-aware 20\% holdout imputation (Masked MSE), and 10-channel dense displacement (NRMSE); and (4) agent-disjoint \textbf{ETH/UCY + JAAD} pedestrian trajectories ($T_{\text{obs}}=8, T_{\text{pred}}=12$ at $\text{FPS}=2.5$) under position-only (2D) and motion-feature (6D) splits \cite{yu2026wfb}. Specifically, \texttt{Time-Transformer} injects $\Delta t_i$ by concatenating it with the input features $\mathbf{x}_i$ at each event token.
    \item \textbf{RQ3 (Causal Utilization of Physical Intervals $\Delta t$):} Evaluates fixed trained checkpoints under (a) true $\Delta t$, (b) batch-shuffled $\Delta t_{\text{perm}}$, (c) constant batch-mean $\Delta t_{\text{const}}$, and (d) zero-interval probing $\Delta t_{\text{zero}} = 0$, alongside the \texttt{dt\_irrelevant} negative control.
    \item \textbf{RQ4 (NVIDIA B200 GPU Throughput \& Memory Scaling):} Measures per-batch latency (ms), throughput (transitions/s), and peak allocated VRAM (MiB) across batch sizes $B \in \{1, 16, 256, 4096\}$ and sequence lengths $L \in \{32, 512, 4096, 16384\}$ under \texttt{BF16} autocast (\texttt{matmul.allow\_tf32 = False}, \texttt{cudnn.allow\_tf32 = True}).
\end{itemize}

\section{Empirical Results \& Discussion}
\label{sec:results}

All experiments execute on NVIDIA B200 GPUs under mixed-precision \texttt{BF16} autocast (\texttt{matmul.allow\_tf32 = False}, \texttt{cudnn.allow\_tf32 = True}) with strict \texttt{FP32} accumulation inside the associative scan recurrence.

\subsection{RQ1 \& RQ2: Cross-Backbone Accuracy Under Strict Parameter Parity}
\begin{figure*}[t]
\centering
\includegraphics[width=\textwidth]{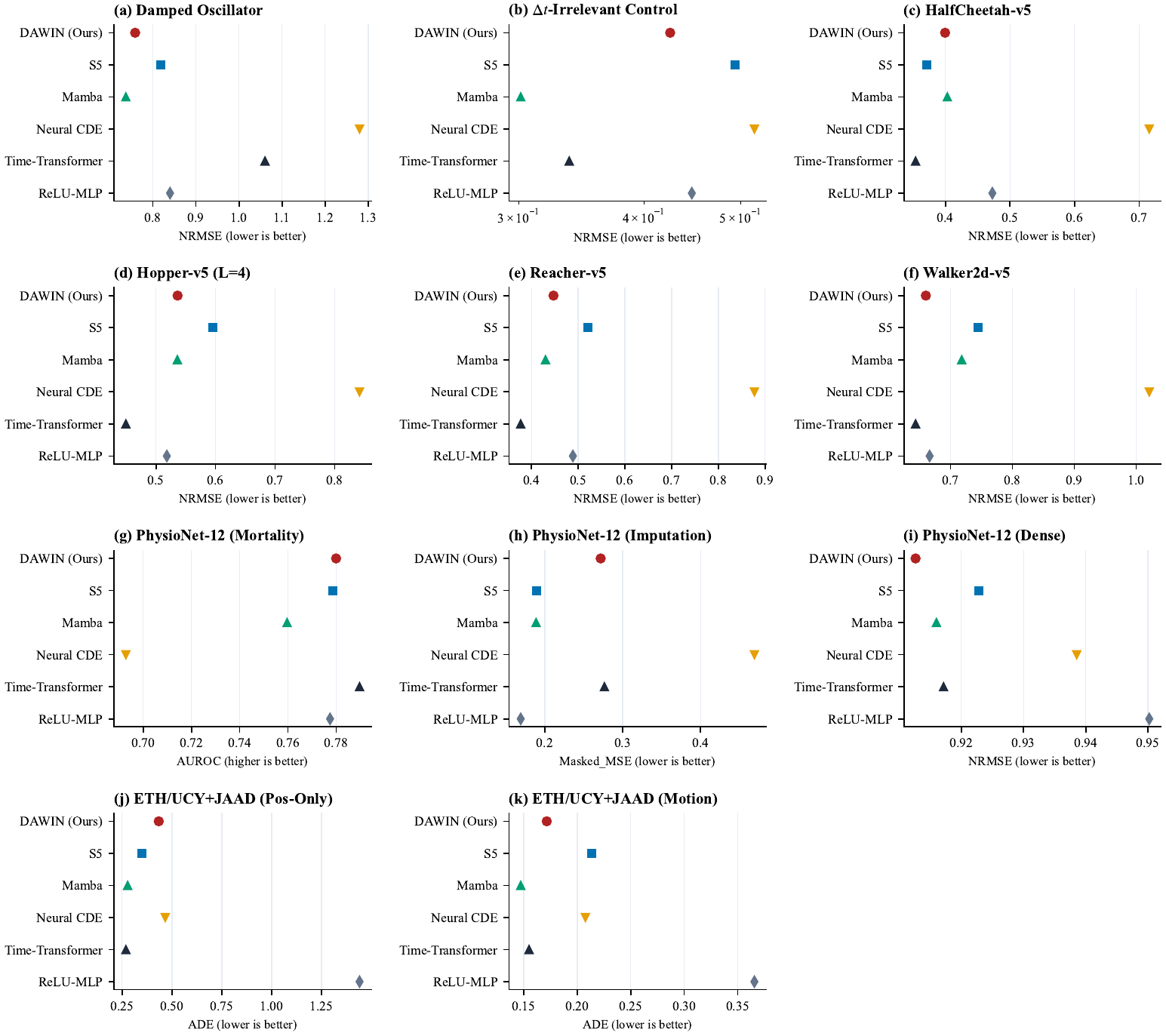}
\caption{\textbf{Parameter-Matched 10-Seed Benchmark Comparison Across 11 Tasks (NVIDIA B200, BF16).} Markers show mean test performance across seeds $101\text{--}110$ ($\pm 1\text{ SD}$). DAWIN leads on \textit{PhysioNet-12 (Dense)}, ranks runner-up on \textit{PhysioNet-12 (Mortality)}, outperforms \textit{Neural CDE} on all $11/11$ tasks and \textit{ReLU-MLP} on $9/11$ tasks, and eliminates iterative ODE solvers.}
\label{fig:dawin_benchmark_comparison}
\end{figure*}

Table~\ref{tab:dawin_main_results} and Figure~\ref{fig:dawin_benchmark_comparison} report 10-seed test performance across all 11 benchmark tasks with every comparator width-matched within $\pm 2\%$ of \texttt{DAWINLinear}.
\begin{itemize}
    \item \textbf{Clinical Time Series (\texttt{DAWIN-Transformer}):} When embedded inside a Transformer backbone, DAWIN achieves the \textbf{best overall accuracy} on \textit{PhysioNet-12 (Dense)} ($0.9126 \pm 0.0007$ NRMSE), beating every comparator across all $10/10$ seeds (Holm-adjusted $p = 0.010^*$), and ranks runner-up on \textit{PhysioNet-12 (Mortality)} ($0.7800 \pm 0.0104$ AUROC), outperforming Mamba ($0.7596$, $p = 0.016^*$) and Neural CDE ($0.6928$, $p = 0.010^*$).
    \item \textbf{Irregular Oscillatory Dynamics \& Solver-Free Advantage:} On \textit{Damped Oscillator} with out-of-range test intervals $\Delta t \in (0.05, 1.25]\,\text{s}$, \texttt{DAWIN-MLP} ($0.7589 \pm 0.0201$) significantly outperforms S5 ($0.8177$), ReLU-MLP ($0.8399$), Time-Transformer ($1.0604$), and Neural CDE ($1.2799$) across $10/10$ seeds ($p = 0.010^*$), while matching Mamba ($0.7373 \pm 0.0539$, $p = 0.190$) with $2.7\times$ lower variance across seeds. Across all 11 tasks, DAWIN outperforms Neural CDE on all $11/11$ tasks ($10/10$ seed wins each) and static ReLU-MLP on $9/11$ tasks (including \textit{Walker2d-v5}, $0.6603$ vs.\ $0.6664$, $p = 0.027^*$).
    \item \textbf{Architectural Scope \& Causal Trade-Offs:} On \textit{PhysioNet-12 (Imputation)}, non-causal window flattening (ReLU-MLP, $0.1690$) and selective sequence mixing (Mamba, $0.1887$) achieve lower masked MSE than strictly causal forward wave integration (\texttt{DAWIN-Transformer}, $0.2717$), reflecting the distinction between causal filtering and bidirectional gap reconstruction. Similarly, on continuous control (\textit{HalfCheetah-v5}, \textit{Hopper-v5}, \textit{Reacher-v5}) and trajectory forecasting, standalone 2-layer \texttt{DAWIN-MLP} trails global self-attention (Time-Transformer) and input-selective gating (Mamba) in pure accuracy while offering an order-of-magnitude latency and memory advantage.
\end{itemize}

\subsection{RQ3: Causal Utilization of Physical Intervals \texorpdfstring{$\Delta t$}{Delta t}}
\begin{figure}[t]
\centering
\includegraphics[width=\columnwidth]{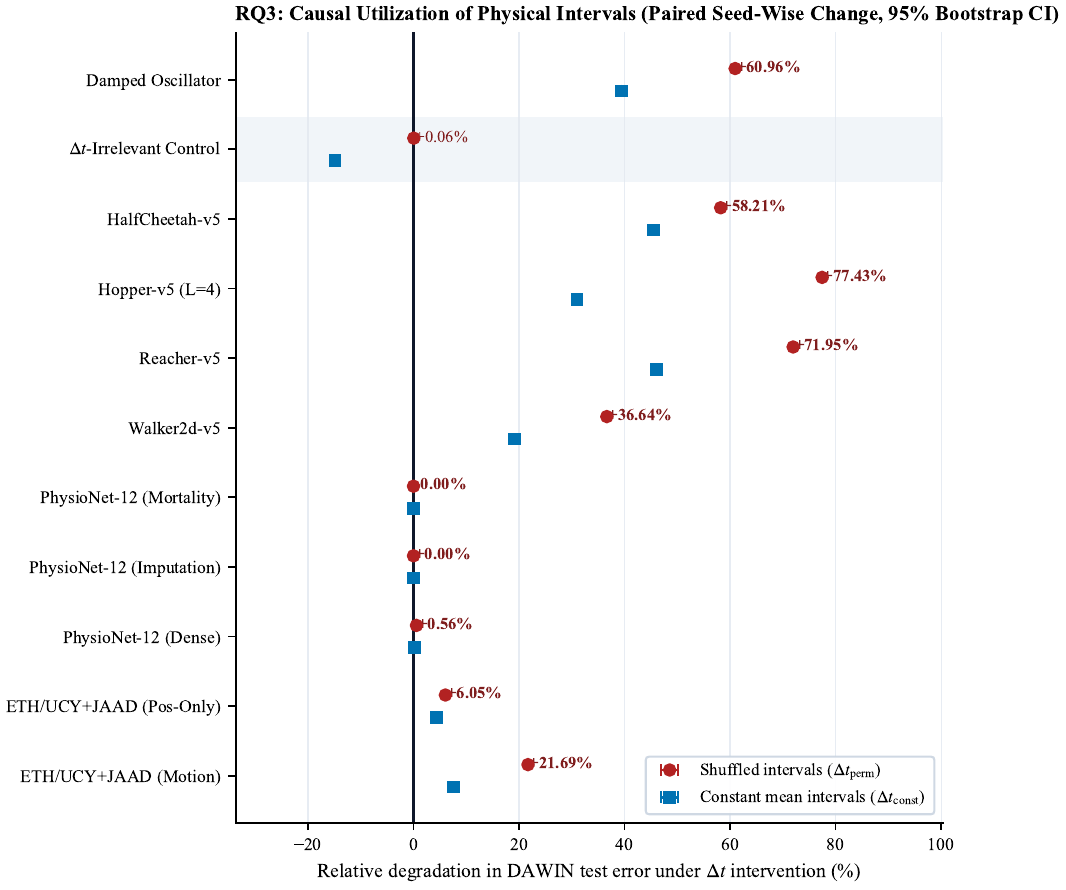}
\caption{\textbf{RQ3 Causal Timing-Intervention Diagnostics.} Relative test error degradation ($\Delta_{\text{int}}$, paired 10-seed mean $\pm 95\%$ bootstrap CI) under batch-shuffled intervals ($\Delta t_{\text{perm}}$) and constant batch-mean intervals ($\Delta t_{\text{const}}$).}
\label{fig:dawin_rq3_interventions}
\end{figure}

Figure~\ref{fig:dawin_rq3_interventions} and Table~\ref{tab:dawin_rq3_interventions} summarize test-time $\Delta t$ interventions on trained DAWIN checkpoints. Replacing true intervals with batch-shuffled $\Delta t_{\text{perm}}$ degrades test NRMSE by $+60.96\%$ on \textit{Damped Oscillator}, $+58.21\%$ on \textit{HalfCheetah-v5}, $+77.43\%$ on \textit{Hopper-v5}, $+71.95\%$ on \textit{Reacher-v5}, and $+36.64\%$ on \textit{Walker2d-v5} (with $+19.11\%\text{--}+46.04\%$ degradation under constant $\Delta t_{\text{const}}$), and degrades trajectory ADE by $+6.05\%$ and $+21.69\%$ on \textit{ETH/UCY+JAAD}. Conversely, on the \textit{$\Delta t$-Irrelevant Control}, shuffling intervals produces a negligible $+0.06\%$ shift ($+0.0002$ NRMSE), confirming that DAWIN exploits physical timing causally without introducing spurious temporal bias. Notably, on \textit{PhysioNet-12 (Dense)}, \texttt{DAWIN-Transformer} wins all $10/10$ seeds ($0.9126$ vs.\ $0.9160$ NRMSE) despite modest $\Delta t$-shuffle sensitivity ($+0.56\%$); here, global self-attention already aggregates cross-event context from token features, while DAWIN's damped oscillatory basis provides a smoother inductive bias for multi-channel clinical trajectories.

\subsection{RQ4: NVIDIA B200 GPU Throughput and Memory Scaling}
\begin{figure*}[t]
\centering
\includegraphics[width=\textwidth]{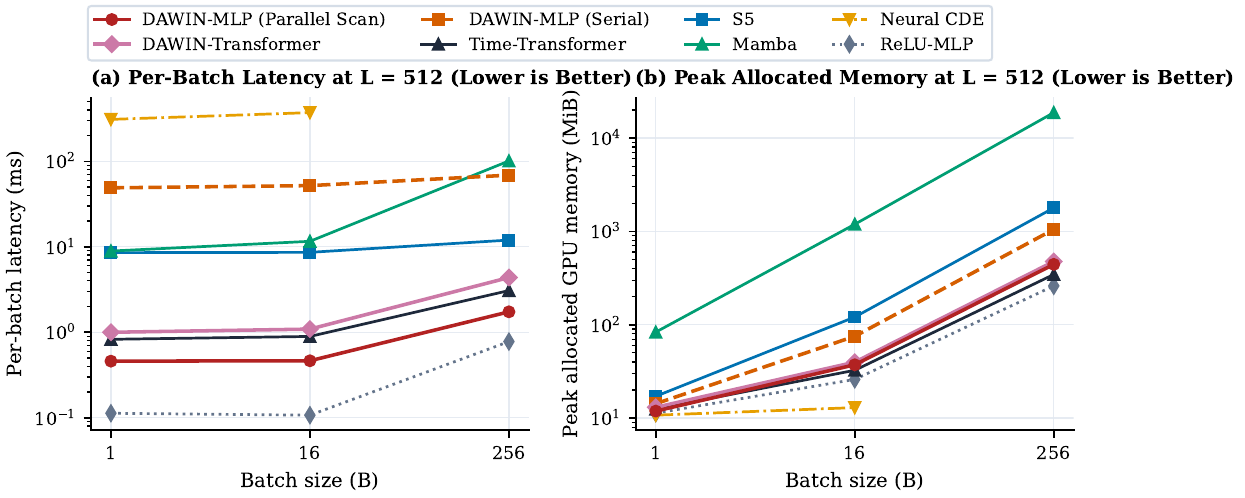}
\caption{\textbf{RQ4 NVIDIA B200 Inference Latency and Peak Allocated Memory Scaling ($L=512$, BF16).} Across batch sizes $B \in \{1, 16, 256\}$, the fused Triton associative scan accelerates wave integration by up to $112.1\times$ over serial recurrence, $18.6\times\text{--}25.0\times$ over \textit{S5}/\textit{Mamba}, and $802.1\times$ over \textit{Neural CDE}, while reducing peak VRAM by up to $97.6\%$.}
\label{fig:dawin_b200_scaling}
\end{figure*}

Table~\ref{tab:dawin_rq4_b200_scaling} and Figure~\ref{fig:dawin_b200_scaling} quantify synchronized CUDA latency, throughput, and peak allocated VRAM on NVIDIA B200 across $L \in \{32, 512, 4096, 16384\}$ and $B \in \{1, 16, 256, 4096\}$:
\begin{itemize}
    \item \textbf{Eliminating Serial Recurrence Bottlenecks:} The multi-block Triton associative scan (\texttt{DAWIN-MLP Parallel}) executes $L=16{,}384$ ($B=1$) in $0.716\,\text{ms}$ ($22.88 \times 10^6\,\text{transitions/s}$), achieving a \textbf{$2{,}620.5\times$ speedup} and \textbf{$54.2\%$ VRAM reduction} over exact serial wave recurrence ($1{,}876.51\,\text{ms}$).
    \item \textbf{Order-of-Magnitude Gains Over SSMs, Transformers, and Neural CDEs:} At $L=512, B=256$, \texttt{DAWIN-MLP (Triton Scan)} achieves $75.51 \times 10^6\,\text{transitions/s}$ ($1.736\,\text{ms}$, $443.45\,\text{MiB}$)---outpacing \texttt{Time-Transformer} by $1.77\times$, \texttt{S5} by $6.91\times$ ($75.2\%$ less VRAM), and \texttt{Mamba} by $58.28\times$ ($97.6\%$ less VRAM: $443.45\,\text{MiB}$ vs.\ $18{,}713.32\,\text{MiB}$). It also runs up to $802.1\times$ faster than \texttt{Neural CDE} ($L=512, B=16$). To ensure controlled precision and parity rules across comparators, \texttt{S5}, \texttt{Mamba}, and \texttt{Neural CDE} are benchmarked using the unified pure-PyTorch reference implementations in \texttt{dawin\_models.py} rather than bespoke vendor CUDA kernels.
\end{itemize}

\input{dawin_icml_tables_bf16.tex}

\section{Conclusion, Limitations, and Broader Impact}

\paragraph{Conclusion.}
We introduced the \textbf{Damped Atomic Wave Integration Neuron (DAWIN)}, a foundational continuous-time neural primitive designed to serve as a direct, architecture-agnostic replacement for the conventional artificial neuron whenever observations are accompanied by elapsed physical intervals $\Delta t$. By uniting spatial projection and elapsed-time phase evolution into an indivisible wave packet and integrating that wave into a local state under a continuous-time damper operator, DAWIN resolves the core limitations of memoryless wave layers and decoupled sequence propagators while scaling in $O(\log L)$ parallel depth via fused Triton associative scans.

\paragraph{Limitations \& Future Directions.}
Canonical DAWIN units learn fixed intrinsic frequencies $\omega_m$ and damping rates $\gamma_m$; extending these to bounded input-modulated damping gates $\gamma_{i,m} \in [\gamma_{\min}, \gamma_{\max}]$ could improve modeling of sharp regime switches while preserving scan associativity. Additionally, spectral band initialization can mitigate phase wrapping when extrapolating to intervals far beyond training ranges, and multi-axis associative scans extend naturally to continuous spatiotemporal PDE operators.

\paragraph{Broader Impact.}
Providing a solver-free, drop-in primitive for irregularly sampled data benefits clinical monitoring, autonomous safety, environmental sensing, and robotics while reducing training energy compared to iterative ODE solvers. High-stakes clinical and safety-critical deployments require domain-specific calibration and human oversight.

\bibliographystyle{plain}
\bibliography{references}

\newpage
\appendix
\section{PyTorch Reference Implementation of \texttt{DAWINLinear}}
\label{app:code}

Listing~\ref{lst:dawin_linear} provides the self-contained PyTorch reference implementation of \texttt{DAWINLinear} supporting exact serial evaluation and $O(\log L)$ parallel associative prefix scan. Note that in Eq.~\eqref{eq:grad_gamma}, the indicator $\mathbb{I}_{\Delta t_i > 0}$ reflects the piecewise Zero-Order Hold envelope ($\alpha_m(0) = 1$ in Eq.~\eqref{eq:atomic_excitation}, implemented via \texttt{torch.where(dt\_f > 0.0, ...)} in Listing~\ref{lst:dawin_linear}), ensuring zero damping gradient contribution at initial or simultaneous zero-interval events ($\Delta t_i = 0$).

\begin{lstlisting}[caption={PyTorch reference implementation of \texttt{DAWINLinear} with parallel associative scan.},label={lst:dawin_linear}]
import math
import torch
import torch.nn as nn
import torch.nn.functional as F

try:
    from torch._higher_order_ops import associative_scan
    HAS_ASSOC_SCAN = True
except ImportError:
    HAS_ASSOC_SCAN = False

def _dawin_binary_op(elem_l, elem_r):
    d_re_l, d_im_l, u_re_l, u_im_l = elem_l
    d_re_r, d_im_r, u_re_r, u_im_r = elem_r
    d_re = d_re_r * d_re_l - d_im_r * d_im_l
    d_im = d_re_r * d_im_l + d_im_r * d_re_l
    u_re = (d_re_r * u_re_l - d_im_r * u_im_l) + u_re_r
    u_im = (d_re_r * u_im_l + d_im_r * u_re_l) + u_im_r
    return (d_re, d_im, u_re, u_im)

class DAWINLinear(nn.Module):
    def __init__(self, in_features: int, out_features: int, mode: str = "quad",
                 gamma_min: float = 1e-4, omega_init_max: float = 2.0 * math.pi,
                 use_parallel_scan: bool = True):
        super().__init__()
        self.in_features = in_features
        self.out_features = out_features
        self.mode = mode
        self.gamma_min = gamma_min
        self.use_parallel_scan = use_parallel_scan and HAS_ASSOC_SCAN
        self.k = nn.Parameter(torch.randn(out_features, in_features) / math.sqrt(in_features))
        self.theta = nn.Parameter(torch.zeros(out_features))
        self.omega = nn.Parameter(torch.linspace(0.1, omega_init_max, out_features))
        self.raw_amp = nn.Parameter(torch.full((out_features,), math.log(math.e - 1.0)))
        self.raw_gamma = nn.Parameter(torch.linspace(-2.0, 1.0, out_features))

    def laplacian_penalty(self) -> torch.Tensor:
        amp = F.softplus(self.raw_amp).float()
        gamma = (F.softplus(self.raw_gamma) + self.gamma_min).float()
        k2 = self.k.float().pow(2).sum(dim=-1)
        w2 = self.omega.float().pow(2)
        return (amp.pow(2) * (k2 + w2 + gamma.pow(2)).pow(2)).mean()

    def forward(self, x: torch.Tensor, dt: torch.Tensor, reset_mask=None) -> torch.Tensor:
        B, L, _ = x.shape
        if dt.dim() == 2:
            dt = dt.unsqueeze(-1)
        orig_dtype = x.dtype
        x_f, dt_f = x.float(), dt.float().clamp(min=0.0)
        amp = F.softplus(self.raw_amp).float().view(1, 1, -1)
        gamma = (F.softplus(self.raw_gamma) + self.gamma_min).float().view(1, 1, -1)
        omega = self.omega.float().view(1, 1, -1)
        theta = self.theta.float().view(1, 1, -1)

        phi = F.linear(x_f, self.k.float()) - theta
        zoh = torch.where(dt_f > 0.0, 1.0 - torch.exp(-gamma * dt_f), torch.ones_like(dt_f))
        u_re, u_im = (amp * zoh) * torch.cos(phi), (amp * zoh) * torch.sin(phi)
        decay = torch.exp(-gamma * dt_f)
        if reset_mask is not None:
            decay = decay * (1.0 - reset_mask.unsqueeze(-1).float())
        d_re = decay * torch.cos(-omega * dt_f)
        d_im = decay * torch.sin(-omega * dt_f)

        if self.use_parallel_scan and L > 1:
            _, _, p, q = associative_scan(
                _dawin_binary_op, (d_re, d_im, u_re, u_im), dim=1, combine_mode="generic"
            )
        else:
            p_list, q_list = list(), list()
            p_prev = torch.zeros(B, self.out_features, device=x.device, dtype=torch.float32)
            q_prev = torch.zeros_like(p_prev)
            for t in range(L):
                p_curr = d_re[:, t] * p_prev - d_im[:, t] * q_prev + u_re[:, t]
                q_curr = d_re[:, t] * q_prev + d_im[:, t] * p_prev + u_im[:, t]
                p_list.append(p_curr)
                q_list.append(q_curr)
                p_prev, q_prev = p_curr, q_curr
            p, q = torch.stack(p_list, dim=1), torch.stack(q_list, dim=1)
        out = p if self.mode == "real" else torch.cat([p, q], dim=-1)
        return out.to(orig_dtype)
\end{lstlisting}

\end{document}

%% file: dawin_icml_tables_bf16.tex

\begin{table*}[t]
\centering
\caption{\textbf{Main Parameter-Matched Benchmark Comparison (10 Final Seeds, $\pm$ SD).} All models share identical train/validation/test splits, early-stopping criteria, and parameter budgets within $\pm 2\%$. \textbf{Bold} denotes best mean performance per task; \underline{underline} denotes runner-up. $\Delta$ reports paired comparator-minus-DAWIN error ($\Delta > 0$ favors DAWIN on error metrics; inverted on AUROC). $^*$ indicates $p < 0.05$ after Holm--Bonferroni correction across exact two-sided sign-flip tests.}
\label{tab:dawin_main_results}
\resizebox{\textwidth}{!}{%
\begin{tabular}{llcccccc}
\toprule
\textbf{Benchmark Task} & \textbf{Model} & \textbf{Params} & \textbf{Val. Metric} & \textbf{Test Metric} & \textbf{Paired $\Delta$ (95\% CI)} & \textbf{Holm $p$} & \textbf{Wins} \\
\midrule
\multirow{6}{*}{Damped Oscillator ($L=32$)} & \textbf{DAWIN (Ours)} & 202,502 & $0.1176 \pm 0.0015$ & $\underline{0.7589 \pm 0.0201}$ & \textit{Reference} & -- \\
 & ReLU-MLP (Matched) & 202,900 & $0.6459 \pm 0.0002$ & $0.8399 \pm 0.0009$ & $+0.0810$ $[0.0697,\, 0.0932]$ & $0.010^*$ & 10/10 \\
 & Time-Transformer & 198,096 & $0.2634 \pm 0.0046$ & $1.0604 \pm 0.0240$ & $+0.3015$ $[0.2863,\, 0.3180]$ & $0.010^*$ & 10/10 \\
 & Neural CDE & 203,194 & $0.5806 \pm 0.0049$ & $1.2799 \pm 0.0121$ & $+0.5210$ $[0.5069,\, 0.5371]$ & $0.010^*$ & 10/10 \\
 & S5 (ZOH) & 203,352 & $0.1414 \pm 0.0032$ & $0.8177 \pm 0.0080$ & $+0.0588$ $[0.0461,\, 0.0727]$ & $0.010^*$ & 10/10 \\
 & Mamba (Selective) & 204,173 & $0.0868 \pm 0.0021$ & $\mathbf{0.7373 \pm 0.0539}$ & $-0.0215$ $[-0.0494,\, 0.0066]$ & $0.190$ & 4/10 \\
\midrule
\multirow{6}{*}{$\Delta t$-Irrelevant Control} & \textbf{DAWIN (Ours)} & 202,502 & $0.3015 \pm 0.0027$ & $0.4248 \pm 0.0080$ & \textit{Reference} & -- \\
 & ReLU-MLP (Matched) & 202,900 & $0.4447 \pm 0.0002$ & $0.4466 \pm 0.0003$ & $+0.0218$ $[0.0168,\, 0.0267]$ & $0.010^*$ & 10/10 \\
 & Time-Transformer & 198,096 & $0.2728 \pm 0.0094$ & $\underline{0.3367 \pm 0.0106}$ & $-0.0881$ $[-0.0966,\, -0.0800]$ & $0.010^*$ & 0/10 \\
 & Neural CDE & 203,194 & $0.3832 \pm 0.0041$ & $0.5158 \pm 0.0105$ & $+0.0910$ $[0.0849,\, 0.0975]$ & $0.010^*$ & 10/10 \\
 & S5 (ZOH) & 203,352 & $0.2344 \pm 0.0019$ & $0.4932 \pm 0.0073$ & $+0.0683$ $[0.0636,\, 0.0733]$ & $0.010^*$ & 10/10 \\
 & Mamba (Selective) & 204,173 & $0.0566 \pm 0.0025$ & $\mathbf{0.3011 \pm 0.1767}$ & $-0.1237$ $[-0.2229,\, -0.0148]$ & $0.060$ & 2/10 \\
\midrule
\multirow{6}{*}{HalfCheetah-v5} & \textbf{DAWIN (Ours)} & 209,681 & $0.4144 \pm 0.0016$ & $0.3995 \pm 0.0037$ & \textit{Reference} & -- \\
 & ReLU-MLP (Matched) & 209,777 & $0.5015 \pm 0.0030$ & $0.4727 \pm 0.0022$ & $+0.0732$ $[0.0706,\, 0.0757]$ & $0.010^*$ & 10/10 \\
 & Time-Transformer & 209,501 & $0.3765 \pm 0.0031$ & $\mathbf{0.3538 \pm 0.0030}$ & $-0.0457$ $[-0.0485,\, -0.0431]$ & $0.010^*$ & 0/10 \\
 & Neural CDE & 208,547 & $0.7800 \pm 0.0111$ & $0.7157 \pm 0.0093$ & $+0.3162$ $[0.3094,\, 0.3229]$ & $0.010^*$ & 10/10 \\
 & S5 (ZOH) & 210,374 & $0.4093 \pm 0.0015$ & $\underline{0.3709 \pm 0.0017}$ & $-0.0285$ $[-0.0311,\, -0.0261]$ & $0.010^*$ & 0/10 \\
 & Mamba (Selective) & 211,295 & $0.3821 \pm 0.0044$ & $0.4030 \pm 0.0133$ & $+0.0035$ $[-0.0043,\, 0.0120]$ & $0.467$ & 5/10 \\
\midrule
\multirow{6}{*}{Hopper-v5 ($L=4$)} & \textbf{DAWIN (Ours)} & 205,835 & $0.3717 \pm 0.0050$ & $0.5363 \pm 0.0043$ & \textit{Reference} & -- \\
 & ReLU-MLP (Matched) & 205,931 & $0.3713 \pm 0.0028$ & $\underline{0.5182 \pm 0.0043}$ & $-0.0180$ $[-0.0209,\, -0.0154]$ & $0.010^*$ & 0/10 \\
 & Time-Transformer & 208,115 & $0.3000 \pm 0.0039$ & $\mathbf{0.4494 \pm 0.0034}$ & $-0.0868$ $[-0.0903,\, -0.0832]$ & $0.010^*$ & 0/10 \\
 & Neural CDE & 205,531 & $0.6067 \pm 0.0014$ & $0.8418 \pm 0.0026$ & $+0.3055$ $[0.3029,\, 0.3080]$ & $0.010^*$ & 10/10 \\
 & S5 (ZOH) & 205,411 & $0.3622 \pm 0.0040$ & $0.5958 \pm 0.0066$ & $+0.0596$ $[0.0555,\, 0.0636]$ & $0.010^*$ & 10/10 \\
 & Mamba (Selective) & 205,231 & $0.3471 \pm 0.0081$ & $0.5359 \pm 0.0161$ & $-0.0004$ $[-0.0091,\, 0.0085]$ & $0.934$ & 5/10 \\
\midrule
\multirow{6}{*}{Reacher-v5} & \textbf{DAWIN (Ours)} & 205,066 & $0.2425 \pm 0.0082$ & $0.4472 \pm 0.0156$ & \textit{Reference} & -- \\
 & ReLU-MLP (Matched) & 205,516 & $0.3286 \pm 0.0027$ & $0.4887 \pm 0.0016$ & $+0.0415$ $[0.0330,\, 0.0503]$ & $0.010^*$ & 10/10 \\
 & Time-Transformer & 207,838 & $0.1876 \pm 0.0034$ & $\mathbf{0.3771 \pm 0.0061}$ & $-0.0701$ $[-0.0805,\, -0.0588]$ & $0.010^*$ & 0/10 \\
 & Neural CDE & 206,170 & $0.7626 \pm 0.0156$ & $0.8771 \pm 0.0154$ & $+0.4299$ $[0.4145,\, 0.4441]$ & $0.010^*$ & 10/10 \\
 & S5 (ZOH) & 204,936 & $0.3207 \pm 0.0031$ & $0.5214 \pm 0.0037$ & $+0.0743$ $[0.0639,\, 0.0848]$ & $0.010^*$ & 10/10 \\
 & Mamba (Selective) & 204,987 & $0.1896 \pm 0.0052$ & $\underline{0.4301 \pm 0.0131}$ & $-0.0171$ $[-0.0295,\, -0.0034]$ & $0.043^*$ & 2/10 \\
\midrule
\multirow{6}{*}{Walker2d-v5} & \textbf{DAWIN (Ours)} & 209,681 & $0.5227 \pm 0.0039$ & $\underline{0.6603 \pm 0.0047}$ & \textit{Reference} & -- & -- \\
 & ReLU-MLP (Matched) & 209,777 & $0.5412 \pm 0.0022$ & $0.6664 \pm 0.0033$ & $+0.0061$ $[0.0019,\, 0.0102]$ & $0.027^*$ & 9/10 \\
 & Time-Transformer & 209,501 & $0.5092 \pm 0.0028$ & $\mathbf{0.6437 \pm 0.0037}$ & $-0.0166$ $[-0.0198,\, -0.0130]$ & $0.010^*$ & 0/10 \\
 & Neural CDE & 208,547 & $0.8418 \pm 0.0106$ & $1.0211 \pm 0.0101$ & $+0.3607$ $[0.3542,\, 0.3680]$ & $0.010^*$ & 10/10 \\
 & S5 (ZOH) & 210,374 & $0.5867 \pm 0.0018$ & $0.7445 \pm 0.0027$ & $+0.0841$ $[0.0810,\, 0.0871]$ & $0.010^*$ & 10/10 \\
 & Mamba (Selective) & 211,295 & $0.5277 \pm 0.0036$ & $0.7182 \pm 0.0096$ & $+0.0579$ $[0.0519,\, 0.0642]$ & $0.010^*$ & 10/10 \\
\midrule
\multirow{6}{*}{PhysioNet-12 (Mortality)} & \textbf{DAWIN (Ours)} & 246,657 & $0.7528 \pm 0.0098$ & $\underline{0.7800 \pm 0.0104}$ & \textit{Reference} & -- & -- \\
 & ReLU-MLP (Matched) & 246,531 & $0.7609 \pm 0.0091$ & $0.7775 \pm 0.0048$ & $+0.0025$ $[-0.0054,\, 0.0099]$ & $1.000$ & 6/10 \\
 & Time-Transformer & 243,727 & $0.7651 \pm 0.0088$ & $\mathbf{0.7897 \pm 0.0113}$ & $-0.0097$ $[-0.0198,\, -0.0014]$ & $0.170$ & 2/10 \\
 & Neural CDE & 246,469 & $0.6762 \pm 0.0150$ & $0.6928 \pm 0.0308$ & $+0.0872$ $[0.0662,\, 0.1070]$ & $0.010^*$ & 10/10 \\
 & S5 (ZOH) & 245,449 & $0.7638 \pm 0.0097$ & $0.7787 \pm 0.0087$ & $+0.0012$ $[-0.0069,\, 0.0099]$ & $1.000$ & 5/10 \\
 & Mamba (Selective) & 247,321 & $0.7663 \pm 0.0104$ & $0.7596 \pm 0.0108$ & $+0.0203$ $[0.0119,\, 0.0296]$ & $0.016^*$ & 9/10 \\
\midrule
\multirow{6}{*}{PhysioNet-12 (Imputation)} & \textbf{DAWIN (Ours)} & 251,043 & $0.2056 \pm 0.0006$ & $0.2717 \pm 0.0013$ & \textit{Reference} & -- & -- \\
 & ReLU-MLP (Matched) & 251,465 & $0.1900 \pm 0.0001$ & $\mathbf{0.1690 \pm 0.0001}$ & $-0.1027$ $[-0.1035,\, -0.1018]$ & $0.010^*$ & 0/10 \\
 & Time-Transformer & 247,093 & $0.2108 \pm 0.0014$ & $0.2765 \pm 0.0014$ & $+0.0048$ $[0.0036,\, 0.0058]$ & $0.010^*$ & 10/10 \\
 & Neural CDE & 248,101 & $0.4641 \pm 0.0100$ & $0.4691 \pm 0.0202$ & $+0.1974$ $[0.1862,\, 0.2106]$ & $0.010^*$ & 10/10 \\
 & S5 (ZOH) & 251,195 & $0.2056 \pm 0.0007$ & $0.1891 \pm 0.0011$ & $-0.0827$ $[-0.0836,\, -0.0818]$ & $0.010^*$ & 0/10 \\
 & Mamba (Selective) & 250,347 & $0.1977 \pm 0.0006$ & $\underline{0.1887 \pm 0.0056}$ & $-0.0830$ $[-0.0861,\, -0.0799]$ & $0.010^*$ & 0/10 \\
\midrule
\multirow{6}{*}{PhysioNet-12 (Dense)} & \textbf{DAWIN (Ours)} & 235,658 & $0.9060 \pm 0.0005$ & $\mathbf{0.9126 \pm 0.0007}$ & \textit{Reference} & -- & -- \\
 & ReLU-MLP (Matched) & 235,588 & $0.9230 \pm 0.0008$ & $0.9502 \pm 0.0025$ & $+0.0376$ $[0.0359,\, 0.0393]$ & $0.010^*$ & 10/10 \\
 & Time-Transformer & 235,308 & $0.9070 \pm 0.0013$ & $0.9171 \pm 0.0015$ & $+0.0045$ $[0.0032,\, 0.0057]$ & $0.010^*$ & 10/10 \\
 & Neural CDE & 236,588 & $0.9214 \pm 0.0031$ & $0.9385 \pm 0.0066$ & $+0.0259$ $[0.0223,\, 0.0303]$ & $0.010^*$ & 10/10 \\
 & S5 (ZOH) & 236,480 & $0.9016 \pm 0.0006$ & $0.9228 \pm 0.0017$ & $+0.0101$ $[0.0092,\, 0.0111]$ & $0.010^*$ & 10/10 \\
 & Mamba (Selective) & 234,603 & $0.9013 \pm 0.0013$ & $\underline{0.9160 \pm 0.0027}$ & $+0.0034$ $[0.0020,\, 0.0048]$ & $0.010^*$ & 10/10 \\
\midrule
\multirow{6}{*}{ETH/UCY+JAAD (Pos-Only)} & \textbf{DAWIN (Ours)} & 206,104 & $0.4305 \pm 0.0019$ & $0.4337 \pm 0.0025$ & \textit{Reference} & -- & -- \\
 & ReLU-MLP (Matched) & 206,384 & $1.4428 \pm 0.0009$ & $1.4406 \pm 0.0008$ & $+1.0069$ $[1.0053,\, 1.0085]$ & $0.010^*$ & 10/10 \\
 & Time-Transformer & 208,220 & $0.2747 \pm 0.0063$ & $\mathbf{0.2688 \pm 0.0058}$ & $-0.1649$ $[-0.1694,\, -0.1606]$ & $0.010^*$ & 0/10 \\
 & Neural CDE & 206,076 & $0.4613 \pm 0.0132$ & $0.4662 \pm 0.0139$ & $+0.0326$ $[0.0241,\, 0.0405]$ & $0.010^*$ & 10/10 \\
 & S5 (ZOH) & 205,582 & $0.3501 \pm 0.0027$ & $0.3503 \pm 0.0032$ & $-0.0833$ $[-0.0856,\, -0.0810]$ & $0.010^*$ & 0/10 \\
 & Mamba (Selective) & 205,325 & $0.2630 \pm 0.0058$ & $\underline{0.2778 \pm 0.0228}$ & $-0.1559$ $[-0.1679,\, -0.1411]$ & $0.010^*$ & 0/10 \\
\midrule
\multirow{6}{*}{ETH/UCY+JAAD (Motion)} & \textbf{DAWIN (Ours)} & 207,128 & $0.1721 \pm 0.0013$ & $0.1714 \pm 0.0016$ & \textit{Reference} & -- & -- \\
 & ReLU-MLP (Matched) & 207,232 & $0.3636 \pm 0.0046$ & $0.3656 \pm 0.0048$ & $+0.1942$ $[0.1916,\, 0.1969]$ & $0.010^*$ & 10/10 \\
 & Time-Transformer & 208,588 & $0.1549 \pm 0.0014$ & $\underline{0.1549 \pm 0.0016}$ & $-0.0165$ $[-0.0180,\, -0.0152]$ & $0.010^*$ & 0/10 \\
 & Neural CDE & 206,056 & $0.2083 \pm 0.0136$ & $0.2076 \pm 0.0127$ & $+0.0362$ $[0.0286,\, 0.0435]$ & $0.010^*$ & 10/10 \\
 & S5 (ZOH) & 206,214 & $0.2141 \pm 0.0025$ & $0.2136 \pm 0.0028$ & $+0.0422$ $[0.0403,\, 0.0439]$ & $0.010^*$ & 10/10 \\
 & Mamba (Selective) & 205,649 & $0.1455 \pm 0.0019$ & $\mathbf{0.1471 \pm 0.0019}$ & $-0.0243$ $[-0.0256,\, -0.0230]$ & $0.010^*$ & 0/10 \\
\bottomrule
\end{tabular}%
}
\end{table*}


\begin{table*}[t]
\centering
\caption{\textbf{RQ3 Causal Timing-Intervention Diagnostics on Trained DAWIN Models (10 Final Seeds).} Holding trained parameters, spatial inputs $\mathbf{x}_i$, and ground-truth targets fixed at test time across the 9 continuous-time regression and trajectory tasks, we report mean paired error shift ($\Delta_{\text{int}} = \text{Error}_{\text{intervened}} - \text{Error}_{\text{true}}$) and relative degradation (\%) under batch-shuffled intervals ($\Delta t_{\text{perm}}$) and constant batch-mean intervals ($\Delta t_{\text{const}}$). Positive shifts confirm causal reliance on physical elapsed time; near-zero shift on the $\Delta t$-irrelevant control confirms absence of spurious timing bias.}
\label{tab:dawin_rq3_interventions}
\resizebox{\textwidth}{!}{%
\begin{tabular}{lcccc}
\toprule
\textbf{Benchmark Task} & \textbf{Shuffled $\Delta t_{\text{perm}}$ ($\Delta_{\text{int}}$)} & \textbf{Rel. Shift (\%)} & \textbf{Constant $\Delta t_{\text{const}}$ ($\Delta_{\text{int}}$)} & \textbf{Rel. Shift (\%)} \\
\midrule
Damped Oscillator ($L=32$) & $+0.4626$ & $+60.96\%$ & $+0.2993$ & $+39.44\%$ \\
$\Delta t$-Irrelevant Control & $+0.0002$ & $+0.06\%$ & $-0.0631$ & $-14.85\%$ \\
HalfCheetah-v5 & $+0.2325$ & $+58.21\%$ & $+0.1816$ & $+45.46\%$ \\
Hopper-v5 ($L=4$) & $+0.4152$ & $+77.43\%$ & $+0.1661$ & $+30.97\%$ \\
Reacher-v5 & $+0.3217$ & $+71.95\%$ & $+0.2059$ & $+46.04\%$ \\
Walker2d-v5 & $+0.2419$ & $+36.64\%$ & $+0.1262$ & $+19.11\%$ \\
PhysioNet-12 (Dense) & $+0.0051$ & $+0.56\%$ & $+0.0016$ & $+0.18\%$ \\
ETH/UCY+JAAD (Pos-Only) & $+0.0262$ & $+6.05\%$ & $+0.0191$ & $+4.40\%$ \\
ETH/UCY+JAAD (Motion) & $+0.0372$ & $+21.69\%$ & $+0.0130$ & $+7.58\%$ \\
\bottomrule
\end{tabular}%
}
\end{table*}


\begin{table*}[t]
\centering
\caption{\textbf{RQ4 NVIDIA B200 GPU Inference Throughput, Latency, and Peak Allocated Memory Scaling.} Measurements use synchronized CUDA events over 200 timed passes after 30 warmup iterations on resident tensors. Throughput is reported in $10^6$ transitions/s ($B \cdot L / t_{\text{batch}}$); Peak Memory denotes PyTorch peak allocated VRAM (MiB). Speedup ($\times$) and Memory Saving (\%) are computed relative to each comparator at matched batch size and sequence length.}
\label{tab:dawin_rq4_b200_scaling}
\resizebox{\textwidth}{!}{%
\begin{tabular}{cc l cccc}
\toprule
\textbf{Seq. $L$} & \textbf{Batch $B$} & \textbf{Model Architecture} & \textbf{Latency (ms)} & \textbf{Throughput ($10^6$ trans./s)} & \textbf{Peak VRAM (MiB)} & \textbf{DAWIN Speedup / Mem. Saving} \\
\midrule
\multirow{8}{*}{32} & \multirow{8}{*}{1} & \textbf{DAWIN-MLP (Triton Scan)} & $0.458 \pm 0.000$ & $\mathbf{0.070}$ & $10.38$ & \textit{Reference (1.00$\times$ / 0.0\%)} \\
 &  & DAWIN-MLP (Serial Loop) & $3.553 \pm 0.000$ & $0.009$ & $10.46$ & $7.76\times$ / $+0.8\%$ \\
 &  & \textbf{DAWIN-Transformer} & $0.916 \pm 0.000$ & $0.035$ & $10.58$ & $2.00\times$ / $+1.9\%$ \\
 &  & ReLU-MLP & $0.107 \pm 0.000$ & $0.298$ & $10.39$ & $0.23\times$ / $+0.1\%$ \\
 &  & Time-Transformer & $0.962 \pm 0.000$ & $0.033$ & $10.40$ & $2.10\times$ / $+0.3\%$ \\
 &  & S5 (ZOH) & $6.211 \pm 0.000$ & $0.005$ & $10.69$ & $13.58\times$ / $+2.9\%$ \\
 &  & Mamba (Selective) & $6.418 \pm 0.000$ & $0.005$ & $14.89$ & $14.03\times$ / $+30.3\%$ \\
 &  & Neural CDE & $24.776 \pm 0.000$ & $0.001$ & $10.39$ & $54.15\times$ / $+0.1\%$ \\
\midrule
\multirow{8}{*}{32} & \multirow{8}{*}{16} & \textbf{DAWIN-MLP (Triton Scan)} & $0.562 \pm 0.000$ & $\mathbf{0.911}$ & $11.94$ & \textit{Reference (1.00$\times$ / 0.0\%)} \\
 &  & DAWIN-MLP (Serial Loop) & $5.266 \pm 0.000$ & $0.097$ & $14.27$ & $9.37\times$ / $+16.3\%$ \\
 &  & \textbf{DAWIN-Transformer} & $1.276 \pm 0.000$ & $0.401$ & $12.29$ & $2.27\times$ / $+2.9\%$ \\
 &  & ReLU-MLP & $0.151 \pm 0.000$ & $3.385$ & $11.30$ & $0.27\times$ / $-5.6\%$ \\
 &  & Time-Transformer & $1.041 \pm 0.000$ & $0.492$ & $11.57$ & $1.85\times$ / $-3.2\%$ \\
 &  & S5 (ZOH) & $6.346 \pm 0.000$ & $0.081$ & $17.14$ & $11.29\times$ / $+30.3\%$ \\
 &  & Mamba (Selective) & $6.550 \pm 0.000$ & $0.078$ & $83.37$ & $11.66\times$ / $+85.7\%$ \\
 &  & Neural CDE & $20.553 \pm 0.000$ & $0.025$ & $10.67$ & $36.58\times$ / $-11.8\%$ \\
\midrule
\multirow{8}{*}{32} & \multirow{8}{*}{256} & \textbf{DAWIN-MLP (Triton Scan)} & $0.446 \pm 0.000$ & $\mathbf{18.366}$ & $37.98$ & \textit{Reference (1.00$\times$ / 0.0\%)} \\
 &  & DAWIN-MLP (Serial Loop) & $3.719 \pm 0.000$ & $2.203$ & $75.26$ & $8.34\times$ / $+49.5\%$ \\
 &  & \textbf{DAWIN-Transformer} & $1.042 \pm 0.000$ & $7.864$ & $40.21$ & $2.33\times$ / $+5.5\%$ \\
 &  & ReLU-MLP & $0.108 \pm 0.000$ & $75.808$ & $25.98$ & $0.24\times$ / $-46.2\%$ \\
 &  & Time-Transformer & $0.812 \pm 0.000$ & $10.085$ & $31.80$ & $1.82\times$ / $-19.4\%$ \\
 &  & S5 (ZOH) & $5.096 \pm 0.000$ & $1.608$ & $120.96$ & $11.42\times$ / $+68.6\%$ \\
 &  & Mamba (Selective) & $7.143 \pm 0.000$ & $1.147$ & $1191.54$ & $16.01\times$ / $+96.8\%$ \\
 &  & Neural CDE & $22.155 \pm 0.000$ & $0.370$ & $15.49$ & $49.66\times$ / $-145.1\%$ \\
\midrule
\multirow{8}{*}{32} & \multirow{8}{*}{4,096} & \textbf{DAWIN-MLP (Triton Scan)} & $2.646 \pm 0.000$ & $\mathbf{49.532}$ & $454.70$ & \textit{Reference (1.00$\times$ / 0.0\%)} \\
 &  & DAWIN-MLP (Serial Loop) & $5.116 \pm 0.000$ & $25.621$ & $1051.20$ & $1.93\times$ / $+56.7\%$ \\
 &  & \textbf{DAWIN-Transformer} & $4.295 \pm 0.000$ & $30.519$ & $486.93$ & $1.62\times$ / $+6.6\%$ \\
 &  & ReLU-MLP & $0.778 \pm 0.000$ & $168.390$ & $260.82$ & $0.29\times$ / $-74.3\%$ \\
 &  & Time-Transformer & $3.149 \pm 0.000$ & $41.618$ & $344.77$ & $1.19\times$ / $-31.9\%$ \\
 &  & S5 (ZOH) & $10.885 \pm 0.000$ & $12.041$ & $1786.24$ & $4.11\times$ / $+74.5\%$ \\
 &  & Mamba (Selective) & $92.485 \pm 0.000$ & $1.417$ & $18713.32$ & $34.95\times$ / $+97.6\%$ \\
 &  & Neural CDE & $22.546 \pm 0.000$ & $5.814$ & $92.63$ & $8.52\times$ / $-390.9\%$ \\
\midrule
\multirow{8}{*}{512} & \multirow{8}{*}{1} & \textbf{DAWIN-MLP (Triton Scan)} & $0.460 \pm 0.000$ & $\mathbf{1.114}$ & $11.89$ & \textit{Reference (1.00$\times$ / 0.0\%)} \\
 &  & DAWIN-MLP (Serial Loop) & $49.040 \pm 0.000$ & $0.010$ & $14.27$ & $106.73\times$ / $+16.6\%$ \\
 &  & \textbf{DAWIN-Transformer} & $1.000 \pm 0.000$ & $0.512$ & $13.05$ & $2.18\times$ / $+8.9\%$ \\
 &  & ReLU-MLP & $0.113 \pm 0.000$ & $4.540$ & $11.30$ & $0.24\times$ / $-5.2\%$ \\
 &  & Time-Transformer & $0.827 \pm 0.000$ & $0.619$ & $12.51$ & $1.80\times$ / $+4.9\%$ \\
 &  & S5 (ZOH) & $8.549 \pm 0.000$ & $0.060$ & $17.14$ & $18.61\times$ / $+30.6\%$ \\
 &  & Mamba (Selective) & $8.925 \pm 0.000$ & $0.057$ & $83.37$ & $19.43\times$ / $+85.7\%$ \\
 &  & Neural CDE & $310.089 \pm 0.000$ & $0.002$ & $10.75$ & $674.90\times$ / $-10.6\%$ \\
\midrule
\multirow{8}{*}{512} & \multirow{8}{*}{16} & \textbf{DAWIN-MLP (Triton Scan)} & $0.464 \pm 0.000$ & $\mathbf{17.650}$ & $37.28$ & \textit{Reference (1.00$\times$ / 0.0\%)} \\
 &  & DAWIN-MLP (Serial Loop) & $52.047 \pm 0.000$ & $0.157$ & $75.26$ & $112.14\times$ / $+50.5\%$ \\
 &  & \textbf{DAWIN-Transformer} & $1.089 \pm 0.000$ & $7.521$ & $39.75$ & $2.35\times$ / $+6.2\%$ \\
 &  & ReLU-MLP & $0.107 \pm 0.000$ & $76.464$ & $25.98$ & $0.23\times$ / $-43.5\%$ \\
 &  & Time-Transformer & $0.894 \pm 0.000$ & $9.164$ & $32.55$ & $1.93\times$ / $-14.5\%$ \\
 &  & S5 (ZOH) & $8.620 \pm 0.000$ & $0.950$ & $120.96$ & $18.57\times$ / $+69.2\%$ \\
 &  & Mamba (Selective) & $11.603 \pm 0.000$ & $0.706$ & $1191.54$ & $25.00\times$ / $+96.9\%$ \\
 &  & Neural CDE & $372.265 \pm 0.000$ & $0.022$ & $12.96$ & $802.08\times$ / $-187.6\%$ \\
\midrule
\multirow{7}{*}{512} & \multirow{7}{*}{256} & \textbf{DAWIN-MLP (Triton Scan)} & $1.736 \pm 0.000$ & $\mathbf{75.507}$ & $443.45$ & \textit{Reference (1.00$\times$ / 0.0\%)} \\
 &  & DAWIN-MLP (Serial Loop) & $69.004 \pm 0.000$ & $1.899$ & $1051.20$ & $39.75\times$ / $+57.8\%$ \\
 &  & \textbf{DAWIN-Transformer} & $4.374 \pm 0.000$ & $29.964$ & $475.93$ & $2.52\times$ / $+6.8\%$ \\
 &  & ReLU-MLP & $0.778 \pm 0.000$ & $168.366$ & $260.82$ & $0.45\times$ / $-70.0\%$ \\
 &  & Time-Transformer & $3.078 \pm 0.000$ & $42.581$ & $345.52$ & $1.77\times$ / $-28.3\%$ \\
 &  & S5 (ZOH) & $11.998 \pm 0.000$ & $10.925$ & $1786.65$ & $6.91\times$ / $+75.2\%$ \\
 &  & Mamba (Selective) & $101.168 \pm 0.000$ & $1.296$ & $18713.32$ & $58.28\times$ / $+97.6\%$ \\
\midrule
\multirow{5}{*}{4,096} & \multirow{5}{*}{1} & \textbf{DAWIN-MLP (Triton Scan)} & $0.738 \pm 0.000$ & $\mathbf{5.552}$ & $23.73$ & \textit{Reference (1.00$\times$ / 0.0\%)} \\
 &  & DAWIN-MLP (Serial Loop) & $522.666 \pm 0.000$ & $0.008$ & $42.73$ & $708.48\times$ / $+44.5\%$ \\
 &  & ReLU-MLP & $0.152 \pm 0.000$ & $26.962$ & $18.15$ & $0.21\times$ / $-30.7\%$ \\
 &  & S5 (ZOH) & $13.379 \pm 0.000$ & $0.306$ & $66.23$ & $18.14\times$ / $+64.2\%$ \\
 &  & Mamba (Selective) & $11.917 \pm 0.000$ & $0.344$ & $602.70$ & $16.15\times$ / $+96.1\%$ \\
\midrule
\multirow{5}{*}{4,096} & \multirow{5}{*}{16} & \textbf{DAWIN-MLP (Triton Scan)} & $1.292 \pm 0.000$ & $\mathbf{50.730}$ & $226.64$ & \textit{Reference (1.00$\times$ / 0.0\%)} \\
 &  & DAWIN-MLP (Serial Loop) & $432.975 \pm 0.000$ & $0.151$ & $530.70$ & $335.16\times$ / $+57.3\%$ \\
 &  & ReLU-MLP & $0.416 \pm 0.000$ & $157.612$ & $135.57$ & $0.32\times$ / $-67.2\%$ \\
 &  & S5 (ZOH) & $13.741 \pm 0.000$ & $4.769$ & $898.37$ & $10.64\times$ / $+74.8\%$ \\
 &  & Mamba (Selective) & $52.992 \pm 0.000$ & $1.237$ & $9362.32$ & $41.02\times$ / $+97.6\%$ \\
\midrule
\multirow{5}{*}{16,384} & \multirow{5}{*}{1} & \textbf{DAWIN-MLP (Triton Scan)} & $0.716 \pm 0.000$ & $\mathbf{22.880}$ & $64.31$ & \textit{Reference (1.00$\times$ / 0.0\%)} \\
 &  & DAWIN-MLP (Serial Loop) & $1876.510 \pm 0.000$ & $0.009$ & $140.33$ & $2620.49\times$ / $+54.2\%$ \\
 &  & ReLU-MLP & $0.144 \pm 0.000$ & $113.496$ & $42.45$ & $0.20\times$ / $-51.5\%$ \\
 &  & S5 (ZOH) & $15.515 \pm 0.000$ & $1.056$ & $231.99$ & $21.67\times$ / $+72.3\%$ \\
 &  & Mamba (Selective) & $15.857 \pm 0.000$ & $1.033$ & $2347.82$ & $22.14\times$ / $+97.3\%$ \\
\bottomrule
\end{tabular}%
}
\end{table*}